\documentclass[conference]{IEEEtran}
\IEEEoverridecommandlockouts
\usepackage{cite}
\usepackage{amsmath,amssymb,amsfonts}
\usepackage{algorithmic}
\usepackage{graphicx}
\usepackage{textcomp}
\usepackage{xcolor}
\usepackage{algorithm}
\usepackage{multirow}
\usepackage[utf8]{inputenc}
\usepackage{url}
\usepackage{color}

\def\BibTeX{{\rm B\kern-.05em{\sc i\kern-.025em b}\kern-.08em
    T\kern-.1667em\lower.7ex\hbox{E}\kern-.125emX}}
\begin{document}

\title{Underwater object detection using Invert Multi-Class Adaboost with deep learning\\}

\author{\IEEEauthorblockN{Long Chen, Zhihua Liu, Lei Tong, Zheheng Jiang, Shengke Wang, Junyu Dong, Huiyu Zhou}}
\maketitle

\begin{abstract}
In recent years, deep learning based methods have achieved promising performance in standard object detection. However, these methods lack sufficient capabilities to handle underwater object detection due to these  challenges: (1) Objects in real applications are usually small and their images are blurry, and (2) images in the underwater datasets and real applications accompany heterogeneous noise. To address these two problems, we first propose a novel neural network architecture, namely Sample-WeIghted hyPEr Network (SWIPENet), for small object detection. SWIPENet consists of high resolution and semantic-rich Hyper Feature Maps which can significantly improve small object detection accuracy. In addition, we propose a novel sample-weighted loss function which can model sample weights for SWIPENet, which uses a novel sample re-weighting algorithm, namely Invert Multi-Class Adaboost (IMA), to reduce the influence of noise on the proposed SWIPENet. Experiments on two underwater robot picking contest datasets URPC2017 and URPC2018 show that the proposed SWIPENet+IMA framework achieves better performance in detection accuracy against several state-of-the-art object detection approaches.
\end{abstract}


\section{Introduction}

Underwater object detection aims to localise and recognise objects in underwater scenes. This research has attracted continuous attention because of its widespread applications in the fields such as oceanography \cite{b1}, underwater navigation \cite{b2} and fish farming \cite{b3}. However, it is still a challenging task due to complicated underwater environments and lighting conditions.

\begin{figure*}[tbp]
\centerline{\includegraphics[width=16cm, height=3.5cm]{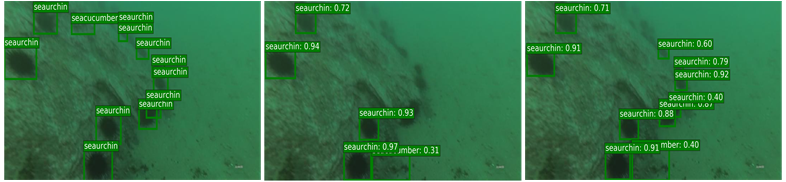}}
\caption{Exemplar images with ground truth annotations (left), results of  Single Shot MultiBox Detector (SSD) (mid) \cite{b25} and our method (right). SSD cannot detect all the small objects while our method outperforms SSD in this case.}
\label{fig:1}
\end{figure*}

Deep learning based object detection systems have demonstrated promising performance in various applications but still felt short in dealing with underwater object detection. This is because, firstly, underwater detection datasets are scarce and the objects in the available underwater datasets and real applications are usually small. Current deep learning based detectors cannot effectively detect small objects (see an example shown in Fig.~\ref{fig:1}). Secondly, the images in the existing underwater datasets and real applications are cluttered. It has been known that in the underwater scenes, wavelength-dependent absorption and scattering \cite{b4} significantly degrade the quality of underwater images. This causes many problems such as visibility loss, weak contrast and color change, which pose numerous challenges to the detection task.

To address these problems, we here propose a deep neural network named Sample-WeIghted hyPEr Network (SWIPENet), which fully takes advantage of multiple Hyper Feature Maps to improve small object detection. Furthermore, we introduce a sample-weighted loss function that cooperates with an Invert Multi-Class Adaboost (IMA) algorithm to reduce the influence of noise on the feature learning of SWIPENet.

The rest of the paper is organised as follows. Section \ref{sec:rela} gives a brief introduction about the related works. Section \ref{sec:net} describes the structure of SWIPENeT and the sample-weighted loss. Section \ref{sec:ima} introduces the Invert Multi-Class Adaboost algorithm. Section \ref{sec:experiments} reports the results of the proposed method on two underwater datasets URPC2017 and URPC2018.

\section{Related Work}
\label{sec:rela}

\subsection{Underwater object detection}
Underwater object detection techniques have been employed in marine ecology studies for many years. Strachan et al. \cite{b5} used color and shape descriptors to recognise fish transported on a conveyor belt, monitored by a digital camera. Spampinato et al. \cite{b6} presented a vision system for detecting, tracking and counting fish in real-time videos, which consist of video texture analysis, object detection and tracking process. However, the above-mentioned methods heavily rely on hand-crafted features, which have limited the representation ability. Choi \cite{b7} applied a foreground detection method to extracting candidate fish windows and used Convolution Neural Networks (CNNs) to classify fish species in the field of view. Ravanbakhsh et al. \cite{b8} compared a deep learning method against the Histogram of Oriented Gridients (HOG)+Support Vector Machine (SVM) method in detecting coral reef fishes, and the experimental results show the superiority of the deep learning methods in underwater object detection. Li et al. \cite{b9} exploited Fast RCNN \cite{b10} to detect and recognise fish species. Li et al. \cite{b11} accelerated fish detection using Faster RCNN \cite{b12}. However, these Fast RCNN methods use features from the last convolution layer of the neural network, which is coarse and cannot be used to effectively detect small objects. In addition, the underwater object detection datasets are extremely scarce that hinders the development of  underwater object detection techniques. Recently, Jian et al. \cite{b13, b14} proposed a underwater dataset for underwater saliency detection, which provides object-level annotations that can be used to evaluate underwater object detection algorithms.

\subsection{Sample re-weighting}
Sample re-weighting can be used to address noisy data problems \cite{b15}. It usually assigns a weight to each sample and then optimises the sample-weighted training loss. For training loss based approaches, we may have two research directions. For example, focal loss \cite{b16} and hard example mining \cite{b17} emphasise samples with higher training loss while self-placed learning \cite{b18} and curriculum learning \cite{b19} encourage learning samples with low loss. The two possible solutions take different assumptions over the training data. The first solution assumes that high loss samples are those to be learned whilst the second one assumes that high loss samples are prone to be disturbing or noisy data. Different from the training loss based methods, Multi-Class Adaboost \cite{b20} re-weights samples according to the classification results. This method focuses on learning misclassified samples through increasing their weights during the iteration. In Section \ref{sec:ima}, we propose a novel detection based sample re-weighting algorithm, namely Invert Multi-Class Adaboost (IMA), to reduce the influence of noise by re-weighting. 

\section{Sample-WeIghted hyPEr Network (SWIPENeT)}
\label{sec:net}

\begin{figure}[tbp]
\centerline{\includegraphics[width=9cm, height=7.5cm]{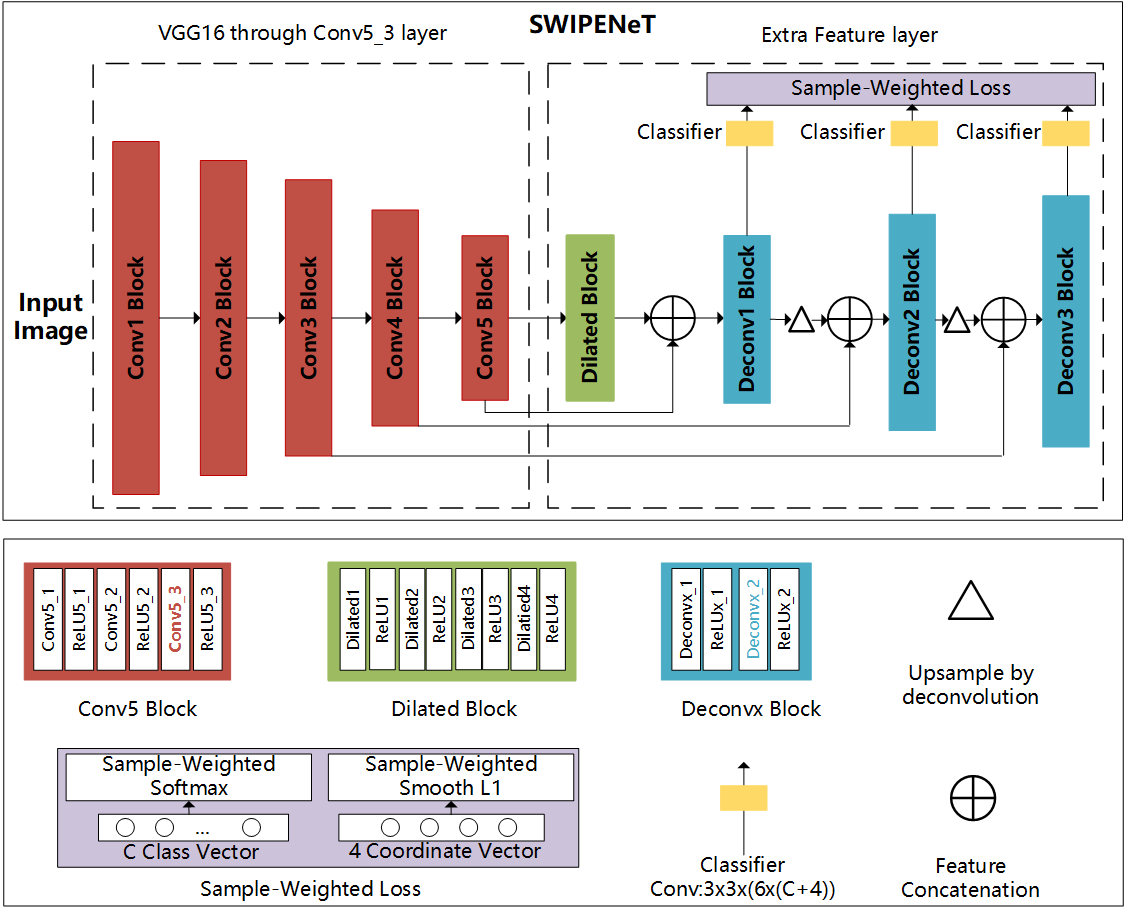}}
\caption{The overview of our proposed SWIPENeT.}
\label{fig:SWIPENeT}
\end{figure}

\subsection{The architecture of our proposed SWIPENeT}\label{AA}
Evidence shows that the down-sampling excises of Convolutional  Neural Network result in strong semantics that lead to the success of many classification tasks. However, this is not enough for the object detection task which not only needs to recognise the category of the object but also spatially locates its position. After we have applied several down-sampling operations, the spatial resolutions of the deep layers are too coarse to handle small object detection.

In this paper, we propose the SWIPENet architecture that includes several high resolution and semantic-rich Hyper Feature Maps inspired by Deconvolutional Single Shot Detector (DSSD) \cite{b21}. DSSD augments a fast down-sampling detection framework SSD \cite{b25} with multiple up-sampling deconvolution layers to increase the resolutions of feature maps. In the DSSD architecture, firstly, multiple down-sampling convolution layers are constructed to extract high semantic feature maps that benefit object classification. After several down-sampling operations, the feature maps are too coarse to provide sufficient information for accurate small object localization, therefore, multiple up-sampling deconvolution layers and skip connection are added to recover the high resolutions of feature maps. However, the detailed information lost by the down-sampling operations cannot be fully recovered even though the resolutions have been recovered. To improve DSSD, we use dilated convolution layers \cite{b22, b23} to obtain strong semantics without losing detailed information that support object localization. Fig.~\ref{fig:SWIPENeT} illustrates the overview of our proposed SWIPENet, which consists of multiple convolution blocks (red), dilated convolution blocks (green), deconvolution blocks (blue) and a novel sample-weigthed loss (gray). The front layers of the SWIPENet are based on the architecture of the standard VGG16 model \cite{b24} (truncated at Conv5\_3 layer). Different from DSSD, we add four dilated convolution layers with ReLU activations to the network, which can obtain large receptive fields without sacrificing the resolutions of the feature maps (large receptive fields lead to strong semantics). We further up-sample feature maps using deconvolution and then use skip connection to pass fine details of low layers to high layers. Finally, we construct multiple Hyper Feature Maps on the deconvolution layers. The prediction of SWIPENet deploys three different deconvolution layers, i.e. Deconv1\_2, Deconv2\_2 and Deconv3\_2 (denoted as Deconvx\_2 in Fig.~\ref{fig:SWIPENeT}), which increase in size progressively and allow us to predict the objects of multiple scales. At each location of the three deconvolution layers, we define 6 default boxes and use a 3$\times$3 convolution kernel to produce $C+1$ class scores ($C$ indicates the number of the object classes and $1$ indicates the background class) and 4 coordinate offsets relative to the original default box shape.

\subsection{Sample-weighted loss}
We propose a novel sample-weighted loss function which can model sample weights in SWIPENeT. The sample-weighted loss enables SWIPENet to focus on learning high weight samples whilst ignoring low weight samples. It cooperates with a novel sample re-weighting algorithm, namely Invert Multi-Class Invert Adaboost, to reduce the influence of possible noise on the SWIPENet by decreasing their weights. Technically speaking, our sample-weighted loss $L$ consists of a sample-weighted Softmax loss $L_{cls}$ for the bounding box classification and a sample-weighted Smooth L1 loss $L_{reg}$ for the bounding box regression (the derivation of the original Softmax loss and Smooth L1 loss can be found in \cite{b25}):

\begin{equation}\small
L=\frac{1}{Num}(\alpha_{1}L_{cls}(pre\_cls,gt\_cls)+\alpha_{2} L_{reg}(pre\_loc,gt\_loc))
\label{formula:1}
\end{equation} 
where
\begin{equation}\small
\begin{split}
L_{cls}=-\sum_{i\in Pos} \sum_{c=1}^{C+1}f(\bar{w}_{i}^{m})gt\_cls_{i}^{c}log(pre\_cls_{i}^{c})\\
-\sum_{i\in Neg} \sum_{c=1}^{C+1}gt\_cls_{i}^{c}log(pre\_cls_{i}^{c})
\label{formula:2}
\end{split}
\end{equation}
\begin{equation}\small
L_{reg}=\sum_{i\in Pos}\sum_{l\in Loc}f(\bar{w}_{i}^{m})SmoothL1(pre\_loc_{i}^{l}-gt\_loc_{j}^{l})
\label{formula:3}
\end{equation}

Following \cite{b25}, SWIPENet trains an object detector using default boxes on several layers. If the Intersection over Union (IoU) between a default box and its most overlapped object is larger than a pre-defined threshold, then the default box is a match to this ground truth object and added to the positive sample set $Pos$. If a default box doesn't match any ground truth object, it will be regarded as negative sample and added to the negative sample set $Neg$. $Num$ is the number of the positive default boxes, $\alpha_{1}$ and $\alpha_{2}$ denote the weight terms of classification loss and regression loss respectively. $C+1$ denotes $C$ object classes plus one background class. $pre\_cls_{i}^{c}$ and $gt\_cls_{i}^{c}$ denote the $c$-th element of the predicted class score and the ground truth class for the $i$-th default box. $pre\_loc_{i}^{l}$ and $gt\_loc_{i}^{l}$ denote the $l$-th element of the predicted coordinate and the ground truth coordinate for the $i$-th positive default box. $Loc=(cx,cy,w,h)$ denotes the object coordinate information that includes the coordinates of center $(cx,cy)$ with width $w$ and height $h$. In our unified detection framework, we train SWIPENet whilst re-weighting each positive sample using IMA. Denote $\bar{w}_{i}^{m}$ as the weight of the $i$-th positive sample learned in IMA in the $m$-th iteration. $f(.)$ is a mapping function that maps weight $\bar{w}_{i}^{m}$ to $f(\bar{w}_{i}^{m})$ which indicates the weight of the $i$-th positive sample used in the sample-weighted loss. We describe the mapping function $f(.)$ in Section \ref{sec:ima} in more details. The sample-weighted loss enables SWIPENet to focus on learning high weight samples and ignore low weight samples.

Sample weights influence the feature learning of SWIPENet through adapting the gradient of the parameters used in back-propagation. Let $w_{cnn}$ be the parameters of SWIPENet, and the gradient of $w_{cnn}$ can be denoted as $\frac{\partial L}{\partial w_{cnn}}$:

\begin{equation}\small
\begin{aligned}
\frac{\partial L}{\partial w_{cnn}}=\frac{\alpha_{1}}{Num}\sum_{i\in Pos}f(\bar{w}_{i}^{m}) \bigtriangledown_{w_{cnn}}^{L_{cls\_pos}^{i}}
+\frac{\alpha_{1}}{Num}\sum_{i\in Neg}\bigtriangledown_{w_{cnn}}^{L_{cls\_neg}^{i}}\\
+\frac{\alpha_{2}}{Num} \sum_{i\in Pos} f(\bar{w}_{i}^{m}) \bigtriangledown_{w_{cnn}}^{L_{loc\_pos}^{i}}
\end{aligned}
\label{formula:4}
\end{equation}

Here, $\bigtriangledown_{w_{cnn}}^{L_{cls\_pos}^{i}}$ and $\bigtriangledown_{w_{cnn}}^{L_{loc\_pos}^{i}}$ indicate the influence of the classification and localisation losses of the $i$-th positive sample on the gradient of the parameters. $\bigtriangledown_{w_{cnn}}^{L_{cls\_neg}^{i}}$ indicates the influence of the $i$-th negative sample's localisation  loss on the gradient of the parameters. The derivation procedure is shown in the Supplementary. It can be seen from (4) that the gradient of the parameters is influenced by the $i$-th positive sample's weight $f(\bar{w}_{i}^{m})$. Specially, $f(\bar{w}_{i}^{m})$ influences the first term and the third term on the right hand side of (4). The smaller the weight is, the smaller gradient in back-propagation is used for the $i$-th sample. For example, if we assign a weight of 1000 and 1 to the same positive sample respectively, then the magnitude of the gradient from the former will be much bigger than that of the gradient from the later. The feature learning of SWIPENeT is dominated by high-weight samples while the feature learning of low-weight samples is ignored.

\section{Invert Multi-Class Adaboost (IMA)}
\label{sec:ima}

\subsection{The overview of IMA}
\begin{figure}[tbp]
\centerline{\includegraphics[width=9cm, height=3.5cm]{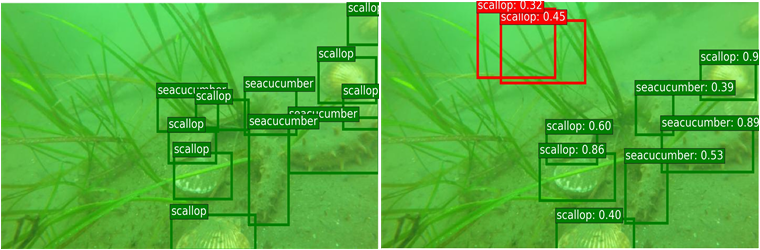}}
\caption{SWIPENeT treats backgrounds as objects on URPC2018. Left is the ground truth annotations and right includes detection result by SWIPENet.}
\label{fig:noisydata}
\end{figure}

SWIPENet possibly misses or incorrectly detectes some objects in the training set, which may be treated as noisy data \cite{b26, b27, b28, b29}. This is because the noisy data are extremely blurry and similar to the background, making them easy to be ignored or detected as the background. If we train the SWIPENet using these noisy data, the performance may be affected. SWIPENet cannot distinguish the background from the objects mainly due to the noise. Fig.~\ref{fig:noisydata} shows exemplar testing images and their incorrect detections by SWIPENet. To handle this problem, we here propose the IMA algorithm inspired by \cite{b30} to reduce the weights of the uncertain objects in order to improve the detection accuracy of SWIPENet. 

IMA is based on Multi-Class Adaboost \cite{b20} which firstly trains multiple base classifiers sequentially and assign a weight value $\alpha_{m}$ according to its error rate $E_{m}$. Then, the samples misclassified by the preceding classifier are assigned a higher weight, allowing the following classifier to focus on learning these samples. Finally, all the weak base classifiers are combined to form an ensemble classifier with corresponding weights. Our IMA also trains M times SWIPENet and then ensemble them into a unified model. Differently, in each training iteration, IMA decreases the weight of the missed objects to reduce the influence of these 'disturbing' samples. The overview of the proposed IMA algorithm can be found in Algorithm 1. $I_{train}$ indicates the training images with the ground truth objects $B=\{b_{1}, b_{2},..., b_{N}\}$, $N$ is the number of the objects in the training set, $b_{j}=(cls,cx,cy,w,h)$ is the annotation of the $j$-th object. We denote $w_{j}^{m}$ as the weight of the $j$-th object in the $m$-th iteration. Each sample's weight is initialised to $\frac{1}{N}$ in the first iteration, i.e. $w_{j}^{1}=\frac{1}{N}, j=1,...,N$.

In the $m$-th iteration, we firstly compute the weights of the positive samples. If the $i$-th positive sample matches the $j$-th object during the training, we assign the $j$-th object's weight $w_{j}^{m}$ as the $i$-th positive sample's weight $\bar{w}_{i}^{m}$, i.e. $\bar{w}_{i}^{m}=w_{j}^{m}$. The mapping function $f(.)$ maps the weight $\bar{w}_{i}^{m}$ to the weight $f(\bar{w}_{i}^{m})$ used in the sample-weighted loss by (10). Secondly, we use $f(\bar{w}_{i}^{m})$ to train the $m$-th SWIPENeT $G_{m}$. Thirdly, we run the $m$-th SWIPENeT on the training set and get the detection set $D_{m}=\{d_{1}, d_{2},..., d_{i}\}$ while $d_{i}=(cls,score,cx,xy,w,h)$ is the result of the $i$-th detected box, including class ($cls$), score ($score$) and coordinates ($cx,cy,w,h$). We compute the $m$-th SWIPENeT's error rate $E_{m}$ based on the percentage of the undetected objects.

\begin{equation}\small
E_{m}=\sum_{j=1}^{N}w_{j}^{m}I(b_{j})/\sum_{j=1}^{N}w_{j}^{m}
\label{formula:5}
\end{equation}
where
\begin{equation}\small
    \displaystyle
    I(b_{j})=\left\{
             \begin{array}{lr}
             0,\,\exists\, d\in D_{m},\, s.t. b_{j}.cls=d.cls \wedge IoU(b_{j},d)\geq \theta\, \\
             1,\,otherwise \\  
             \end{array}
\right.
\label{formula:6}
\end{equation}

If there exists a detection $d$ which belongs to the same class as the $j$-th ground truth object $b_{j}$ (i.e. $b_{j}.cls=d.cls$) and the Intersection over Union (IoU) between the detection and the $j$-th object is larger than $\theta$ (0.5 here), we set $I(b_{j})=0$ which indicates the $j$-th object has been detected, and $I(b_{j})=1$ indicates the undetected. Fourthly, we compute the $m$-th SWIPENeT's weight $\alpha_{m}$ in the final ensemble model. $C$ is the number of the object classes.
\begin{equation}\small
\alpha_{m}=log\frac{1-E_{m}}{E_{m}}+log(C-1)
\label{formula:7}
\end{equation}

Finally, we update each object's weight $w_{j}^{m}$. Different from Multi-Class Adaboost, we decrease the weights of the undetected objects by (8). $z_{m}$ is a normalization constant. The iteration repeats again till all $M$ SWIPENeT have been trained.
\begin{equation}\small
w_{j}^{m} \leftarrow \frac{w_{j}^{m}}{z_{m}}exp(\alpha _{m}(1-I(b_{j})))
\label{formula:8}
\end{equation}

In the testing stage, we first run all $M$ SWIPENeT on the testing set $I_{test}$ and get $M$ detection set $D_m=G_{m}(I_{test}), m=1,2,...,M$.
Afterwards, we re-score each detection $d$ in $D_{m}$ according to $\alpha_{m}$, i.e. $d.score=\alpha_{m} d.score$.

Finally, we combine all the detections and apply Non-Maximum Suppression \cite{b31} to removing the overlapped detections and generating the final detections by (9).

\begin{equation}\small
D = NonMaximumSuppression(\bigcup_{m=1}^{M} D_{m})
\label{formula:9}
\end{equation}

\begin{algorithm}
\caption{SWIPENeT with Invert Multi-Class Adaboost}
\label{alg:1}
\textbf{Input}: Training images $I_{train}$ with ground truth objects $B=\{b_{1},...,b_{N}\}$, testing images $I_{test}$.\\
\textbf{Output}: Detection results $D$.
\begin{algorithmic}[1] %[1] enables line numbers
\STATE Initialize the object weights $w_{j}^{1}=\frac{1}{N}, j=1,...,N$.
\FOR{$m=1$ to $M$}
\STATE $\bullet$  Compute the weights of positive samples using (10).\\
\STATE $\bullet$  Train the $m$-th SWIPENeT $G_{m}$ using (1)-(3).\\
\STATE $\bullet$ Compute the $m$-th SWIPENeT's error rate $E_{m}$ using (5)-(6).
\STATE $\bullet$ Compute the $m$-th SWIPENeT's weight $\alpha_{m}$ in the ensemble model using (7).
\STATE $\bullet$ Decrease the weights of undetected objects and increase the weights of detected objects using (8).
\ENDFOR
\STATE  Get the final detections $D$ using (9).
\STATE \textbf{return} Detections results $D$
\end{algorithmic}
\end{algorithm}

\subsection{The mapping function f(.)}
We define two weights for the $i$-th positive sample, i.e. IMA weight $\bar{w}_{i}^{m}$ which is learned in IMA and $f(\bar{w}_{i}^{m})$ which indicates the weight used in the sample-weighted loss. In IMA, the initial IMA weight of each positive sample is $\frac{1}{N}$ ($N$ is the number of the objects in the training data) and the initial weight of each positive sample used in the sample-weighted loss is 1. Hence, in the sample-weighted loss function, the positive samples' weights are $N$ times their IMA weights. Intuitively, we can define a linear mapping function $f(.)$ to map the IMA weight to the weight used in the sample-weighted loss. Here, we first assign the weight of the $j$-th object as the weight of the $i$-th default box if they match, we denote the weights of the $j$-th object and $i$-th default box as $w_{j}^{m}$ and $\bar{w}_{i}^{m}$, respectively. We have $\bar{w}_{i}^{m}=w_{j}^{m}$. Then, we map IMA weight $\bar{w}_{i}^{m}$ to $f(\bar{w}_{i}^{m})$ where the sample-weighted loss can be used by a linear mapping function stated in (10).

\begin{equation}\small
f(\bar{w}_{i}^{m}) = N*\bar{w}_{i}^{m}, 0<\bar{w}_{i}^{m}<1
\label{formula:10}
\end{equation}

\section{Experiments on URPC2017 and URPC2018}
\label{sec:experiments}

We evaluate our approach on two underwater datasets URPC2017 and URPC2018 from the Underwater Robot Picking Contest. The URPC2017 dataset has 3 object categories, including seacucumber, seaurchin and scallop. There are 18,982 training images and 983 testing images. The URPC2018 dataset has 4 object categories, including seacucumber, seaurchin, scallop and starfish. There are 2,897 images in the training set, but the testing set is not publicly available. We randomly split the training set of URPC2018 into a training set of 1,999 images and a testing set of 898 images. Both two datasets provide underwater images and box level annotations (detailed descriptions are provided in the Supplementary). In this section, we firstly analyse our method using ablation studies in Subsection \ref{abls}. Then, we compare our method against the other state of the art detection frameworks, including SSD \cite{b25}, YOLOv3 \cite{b32} and Faster RCNN \cite{b12} shown in Subsection \ref{contrast}.

\subsection{Ablation studies}\label{abls}
We study the role of several components in our SWIPENet in this section, including dilated convolution layers, skip connection and IMA.

\textbf{Implementation details.} To investigate the influence of dilated convolution and skip connection on our SWIPENet, we design two networks for the comparisons. All the networks in the ablation studies are trained with the Adam optimisation algorithm on a single NVIDIA Tesla P100 GPU with a 16 GB memory. We use an image scale of 512x512 as the input for both training and testing. The source code is developed upon Keras and will be published\footnote{\url{https://github.com/LongChenCV/SWIPENet}}. For URPC2017, the batch-size is 16, the learning rate is 0.0001, our models often diverge when we  use a high learning rate due to unstable gradients, and all the networks achieve the best performance after running 120 epochs. For URPC2018, the batch-size is 16, our models converges quickly when we use a high learning rate 0.001. All the networks achieve the best performance after running 80 epochs.

\begin{table}[tbp]
\begin{center}
\caption{Ablation studies on URPC2017 and URPC2018. Skip indicates skip connection, and Dilation indicates dilated convolution layer. mAP indicates mean Average Precision(\%).}
\begin{tabular}{|c|c|c|c|c|}
\hline
Dataset & Network & Skip & Dilation & mAP\\
\hline
\multirow{3}{*}{URPC2017} & UWNet1 & & \checkmark & 40.4\\ 
& UWNet2 & & & 38.3\\
& SWIPENeT & \checkmark & \checkmark & 42.1\\
\hline
\multirow{3}{*}{URPC2018} & UWNet1 & & \checkmark & 61.2\\ 
& UWNet2 & & & 58.1\\
& SWIPENeT & \checkmark & \checkmark & 62.2\\ 
\hline
\end{tabular}
\end{center}
\label{tab:abla1}
\end{table} 

\textbf{Ablation studies on skip connection and dilated convolution layer.} To investigate the influence of skip connection, we design the first baseline network UWNet1 which has the same structure as SWIPENet except that it does not contain skip connection between the low and high layers. The second network UWNet2 replaces the four dilated convolution layers in UWNet1 with convolution layers to learn the influence of dilated convolution. Table I shows the performance comparison of different networks on URPC2017 and URPC2018. SWIPENet performs 1.7\% and 1.0\% better than UWNet1 on the two datasets respectively. The gains come from the skip connection which passes fine detailed information of the lower layers such as object boundary to the high layers that are important for object localisation. Compared to UWNet2, UWNet1 performs 2.1\% and 3.1\% improvement because the dilated convolution in UWNet1 brings much semantic information to the high layers which enhances the classification ability.

\begin{table}[tbp]
\caption{The performance of SWIPENeT (mAP(\%)) in each iteration of IMA.}
\begin{center}
\begin{tabular}{|c|c|c|c|}
\hline
Dataset & IMA iteration & Single & Ensemble\\
\hline
\multirow{5}{*}{URPC2017} & 1 & 42.1 & -\\ 
& 2 & 44.2 & 45.0\\
& 3 & 45.3 & 46.3\\
& 4 & 40.5 & 45.3\\
& 5 & 37.2 & 44.2\\
\hline
\multirow{5}{*}{URPC2018} & 1 & 62.2 & -\\ 
& 2 & 63.3 & 64.5\\
& 3 & 62.4 & 64.0\\
& 4 & 61.2 & 62.8\\
& 5 & 59.3 & 62.1\\
\hline
\end{tabular}
\end{center}
\label{tab:abla2}
\end{table} 

\textbf{Ablation studies on IMA.} Table \ref{tab:abla2} shows the performance of the single model and the ensemble model after each iteration. The ensemble model has better performance on the two datasets. By reducing the influence of noise, IMA gives SWIPENet 4.2\% and 2.3\% improvement on the two datasets respectively. The single model and the ensemble model both perform best in the 3rd iteration on URPC2017 and in the 2nd iteration on URPC2018. However, the performance of the two models goes down as most of the detected objects are continuously up-weighted with the increasing of the iteration number where SWIPENet over-fits with the high-weight objects. According to the experimental results in Table \ref{tab:abla2}, we set the number of iterations as 3 on URPC2017 and 2 on URPC2018.

\begin{figure}[tbp]
\centerline{\includegraphics[width=8cm, height=7.5cm]{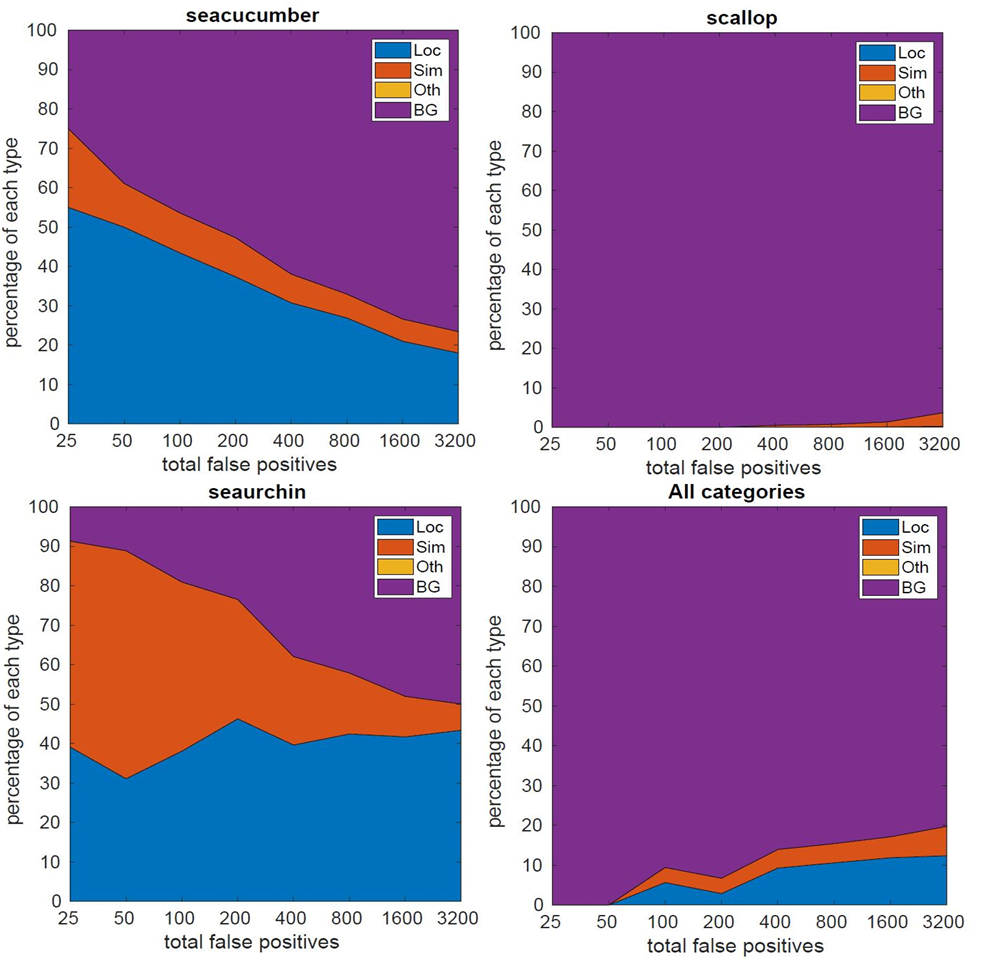}}
\caption{The distribution of top-ranked false positive types for each category and all categories on URPC2017. The false positive types include localisation error (Loc), confusion with similar categories (Sim), with others (Oth), or with background (BG).}
\label{fig:errors}
\end{figure}

We use the detection analysis tool of \cite{b33} to analyse the false positives of SWIPENeT without IMA. Fig.~\ref{fig:errors} shows the distribution of the top-ranked false positive types for each category of the URPC2017 testing set. SWIPENeT cannot well distinguish objects with complex background and localise objects accurately due to the noise in the data.

\subsection{Comparison with state-of-the-art detection frameworks}\label{contrast}
\begin{table}[tbp]\scriptsize  
\caption{Comparison with the state-of-arts on URPC2017.}
\begin{center}
\begin{tabular}{|c|c|c|c|c|c|}
\hline
Methods & Backbone & seacucumber & seaurchin & scallop & mAP\\
\hline
SSD300 & VGG16 & 28.1 & 51.3 & 21.2 & 33.5\\
SSD512 & VGG16 & 38.4 & 52.9 & 15.7 & 35.7\\
YOLOv3 & DarkNet53 & 28.4 & 50.3 & 22.4 & 33.7\\
Faster RCNN & VGG16 & 27.2 & 45.0 & 31.9 & 34.7\\
Faster RCNN & ResNet50 & 31.0 & 41.4 & 33.5 & 35.3\\
Faster RCNN & ResNet101 & 26.2 & 47.7 & 32.5 & 35.5\\
\hline
OurFirstSingle & SWIPENeT & 43.6 & 51.3 & 31.2 & 42.1\\
OurBestSingle & SWIPENeT & \textbf{45.0} & 49.7 & 41.3 & 45.3\\
OurEnsemble & SWIPENeT & 44.4 & \textbf{52.4} & \textbf{42.1} & \textbf{46.3}\\
\hline
\end{tabular}
\end{center}
\label{tab:sot1}
\end{table}

\begin{table}[tbp]\tiny
\caption{Comparison with the state-of-arts on URPC2018.}
\begin{center}
\begin{tabular}{|c|c|c|c|c|c|c|}
\hline
Methods & Backbone & seacucumber & seaurchin & scallop & starfish & mAP\\
\hline
SSD300 & VGG16 & 38.5 & 83.0 & 30.8 & 75.1 & 56.9\\
SSD512 & VGG16 & 44.2 & \textbf{84.4} & 35.8 & 78.1 & 60.6\\
YOLOv3 & DarkNet53 & 35.7 & 83.0 & 34.0 & 77.9 & 57.7\\
Faster RCNN & VGG16 & 43.3 & 83.0 & 32.0 & 74.5 & 58.2\\
Faster RCNN & ResNet50 & 41.1 & 83.2 & 34.5 & 77.2 & 59.0\\
Faster RCNN & ResNet101 & 44.3 & 82.5 & 34.7 & 77.5 & 59.8\\
\hline
OurFirstSingle & SWIPENeT & 46.4 & 84.0 & 40.2 & 78.2 & 62.2\\
OurBestSingle & SWIPENeT & 50.3 & 83.7 & 39.8 & \textbf{79.4} & 63.3\\
OurEnsemble & SWIPENeT & \textbf{52.8} & 84.1 & \textbf{42.9} & 78.0 & \textbf{64.5}\\
\hline
\end{tabular}
\end{center}
\label{tab:sot2}
\end{table}

In this section, we compare our method with other state-of-the-art detection frameworks, including SSD \cite{b16}, YOLOv3 \cite{b19} and Faster RCNN \cite{b22}.

\textbf{Implementation details.} For SSD, we use VGG16 \cite{b19} as the backbone network and conduct experiments on two SSD with different input sizes, i.e. SSD300 and SSD512. For Faster RCNN, we use three backbone networks including VGG16, ResNet50 \cite{b34} and ResNet101 \cite{b34}. For YOLOv3, we use its original DarkNet53 network. When we train SWIPENeT, the parameter setting is the same as that used in the ablation studies.

Tables \ref{tab:sot1} and \ref{tab:sot2} show experimental results on URPC2017 and URPC2018. On URPC2017, SSD512 achieves 35.7 mAP, which improves 2.2\% over SSD300. The gain comes from the increase of the input size. Faster RCNN with ResNet101 performs better than Faster RCNN with ResNet50 and VGG16, where the ResNet-101 plays a critical role. In addition, SSD512 achieves better performance than Faster RCNN, even though Faster RCNN uses ResNet101 as the backbone network, which has better performance than VGG16. It is because SSD512 detects multi-scale objects on different layers, and performs better than Faster RCNN on small object detection. OurFirstSingle, the SWIPENeT trained in the first iteration of IMA, outperforms all the state-of-the-arts by a large margin (above 6.4\%) on URPC2017, demonstrating the superiority of our SWIPENeT in detecting small objects. OurBestSingle, the best performing single SWIPENeT, improves 3.2\% over OurFirstSingle. We ensemble all the SWIPENeTs into OurEnsemble, and this further improves the results to 46.3 mAP. The gain comes from the ensemble model. In addition, both OurBestSingle and OurEnsemble surpass the best result in the URPC2017 competition, and the official leaderboard of the URPC2017 competition is shown in the Supplementary. Fig.~\ref{fig:roc17} shows the Precision/Recall curves of different methods on URPC2017. OurEnsemble (black curve) performs best on the seaurchin and scallop categories, and OurBestSingle (red curve) performs better than OurEnsemble on the seacucumber category, which indicates model ensembling may not bring performance improvement for a single object category.

\begin{figure}[tbp]
\centerline{\includegraphics[width=9cm, height=3.2cm]{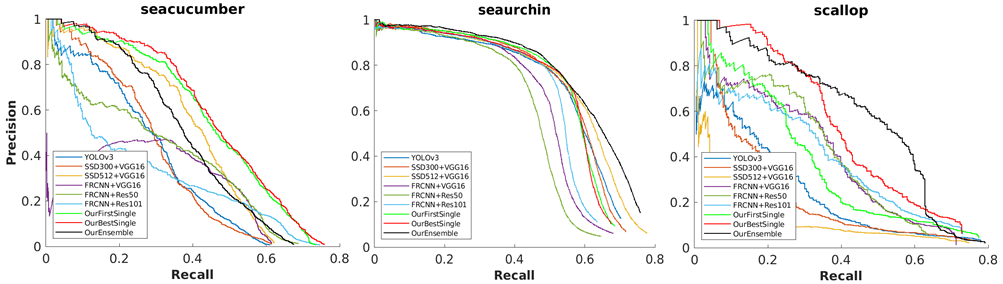}}
\caption{Precision/Recall curves of different methods on URPC2017.}
\label{fig:roc17}
\end{figure}

\begin{figure}[tbp]
\centerline{\includegraphics[width=9cm, height=7.5cm]{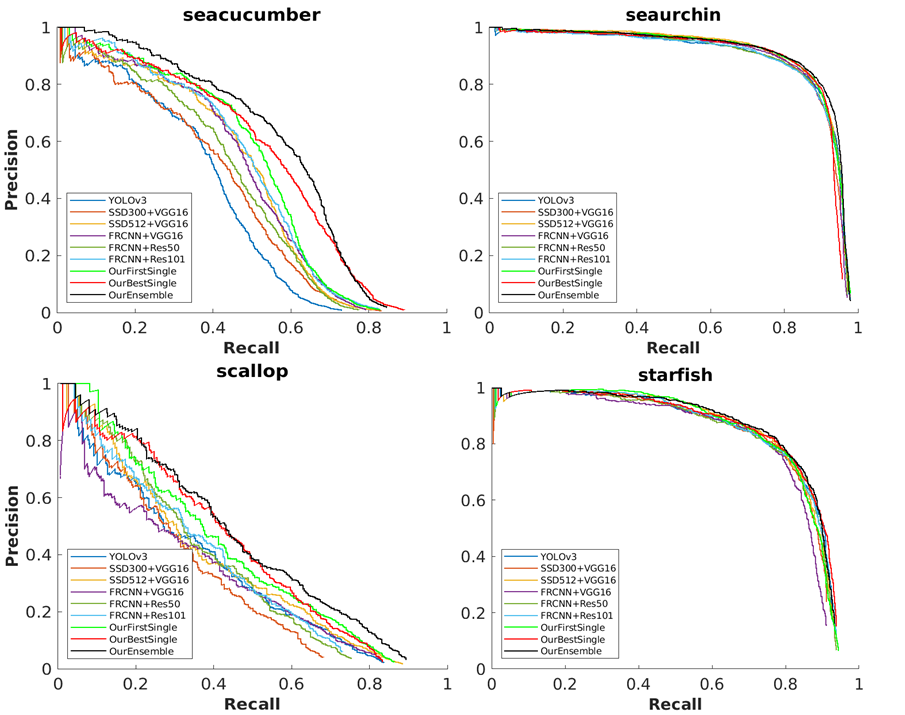}}
\caption{Precision/Recall curves of different methods on URPC2018.}
\label{fig:roc18}
\end{figure}

OurFirstSingle achieves 62.2 mAP on URPC2018 and outperforms the other state-of-the-arts. OurBestSingle improves 1.1\% over OurFirstSingle, and OurEnsemble achieves the best performance, 64.5 mAP. OurEnsemle outperforms all the other state-of-the-art methods by a large margin (above 3.9\%), demonstrating its superiority in detecting small objects and handling noisy data. Fig.~\ref{fig:roc18} shows the Precision/Recall curves of different methods on URPC2018. OurEnsemble performs the best on the seacucumber and scallop categories. All the methods achieves higher accuracy on the seaurchin and starfish categories than on the seacucumber and scallop categories.

\section{Conclusion}
In this paper, we proposed a neural network architecture, called Sample-WeIghted hyPEr Network (SWIPENet), for small underwater object detection. Moreover, a sample re-weighting algorithm named Invert Multi-Class Adaboost (IMA) had been presented to solve the noise issue. Our proposed method achieved state-of-the-art performance on challenging datasets, but its time complexity is M times higher than a single model since it is an ensemble of M deep neural networks. Hence, in future work, reducing the computational complexity of our proposed method is of vital importance. In addition, current deep models introduce attention mechanisms and novel loss to solve the issues of noise and small objects detection, which provide us insightful ideas to develop our SWIPENet.

\section*{Acknowledgment}
Thanks for National Natural Science Foundation of China and Dalian Municipal People's Government providing the underwater object detection datasets for research purposes. This project of underwater object detection is supported by China Scholarship Council.

\section{Supplementary}
\subsection{The derivation of the sample-weighted loss}
Our sample-weighted loss $L$ consists of a sample-weighted Softmax loss for the bounding box classification ($L_{cls}$) and a sample-weighted Smooth L1 loss for the bounding box regression ($L_{reg}$):
\begin{equation}\small
L=\frac{1}{Num}(\alpha_{1}L_{cls}(pre\_cls,gt\_cls)+\alpha_{2} L_{reg}(pre\_loc,gt\_loc))
\label{formula:1}
\end{equation} 
where
\begin{equation}\small
\begin{split}
L_{cls}=-\sum_{i\in Pos} \sum_{c=1}^{C+1}f(\bar{w}_{i}^{m})gt\_cls_{i}^{c}log(pre\_cls_{i}^{c})\\
-\sum_{i\in Neg} \sum_{c=1}^{C+1}gt\_cls_{i}^{c}log(pre\_cls_{i}^{c})
\label{formula:2}
\end{split}
\end{equation}
\begin{equation}\small
L_{reg}=\sum_{i\in Pos}\sum_{l\in Loc}f(\bar{w}_{i}^{m})SmoothL1(pre\_loc_{i}^{l}-gt\_loc_{j}^{l})
\label{formula:3}
\end{equation}
Then we obtain the partial derivative of the network parameters w.r.t. sample-weighted loss.

\begin{equation}\footnotesize
\begin{aligned}
\frac{\partial L}{\partial w_{cnn}}=\frac{\alpha_{1}}{Num} \frac{\partial L_{cls}(pre\_cls,gt\_cls)}{\partial w_{cnn}}+\frac{\alpha_{2}}{Num} \frac{\partial L_{reg}(pre\_loc,gt\_loc)}{\partial w_{cnn}}
\\=\frac{\alpha_{1}}{Num} \sum_{i\in Pos} f(\bar{w}_{i}^{m}) \sum_{c=1}^{C+1} \frac{\partial gt\_{cls}_{i}^{c}log \frac{1}{pre\_cls_{i}^{c}}}{\partial w_{cnn}} 
\\+\frac{\alpha_{1}}{Num}\sum_{i\in Neg} \sum_{c=1}^{C+1} \frac{\partial gt\_{cls}_{i}^{c}log \frac{1}{pre\_cls_{i}^{c}}}{\partial w_{cnn}} 
\\+\frac{\alpha_{2}}{Num}\sum_{i\in Pos} f(\bar{w}_{i}^{m}) \frac{\partial \sum_{l\in Loc} SmoothL1(pre\_loc_{i}^{l}-gt\_loc_{i}^{l})}{\partial w_{cnn}}
\end{aligned}
\end{equation}

Let 
\begin{equation}\small
\bigtriangledown_{w_{cnn}}^{L_{cls\_pos}^{i}}=\sum_{c=1}^{C+1} \frac{\partial gt\_{cls}_{i}^{c}log \frac{1}{pre\_cls_{i}^{c}}}{\partial w_{cnn}} 
\end{equation}
\begin{equation}\small
\bigtriangledown_{w_{cnn}}^{L_{cls\_neg}^{i}}=\sum_{c=1}^{C+1} \frac{\partial gt\_{cls}_{i}^{c}log \frac{1}{pre\_cls_{i}^{c}}}{\partial w_{cnn}} 
\end{equation}
\begin{equation}\small
\bigtriangledown_{w_{cnn}}^{L_{reg\_pos}^{i}}=\frac{\partial \sum_{l\in Loc} SmoothL1(pre\_loc_{i}^{l}-gt\_loc_{i}^{l})}{\partial w_{cnn}}
\end{equation}
Here, $\bigtriangledown_{w_{cnn}}^{L_{cls\_pos}^{i}}$ and $\bigtriangledown_{w_{cnn}}^{L_{reg\_pos}^{i}}$ denote the partial derivatives of the $i$-th positive sample on classification loss and regression loss respectively. $\bigtriangledown_{w_{cnn}}^{L_{cls\_neg}^{i}}$ is the partial derivative of the $i$-th negative sample on the classification loss. Then, we have 
\begin{equation}\small
\begin{split}
\frac{\partial L}{\partial w_{cnn}}=\frac{\alpha_{1}}{Num}\sum_{i\in Pos}f(\bar{w}_{i}^{m}) \bigtriangledown_{w_{cnn}}^{L_{cls\_pos}^{i}}
+\frac{\alpha_{1}}{Num}\sum_{i\in Neg}\bigtriangledown_{w_{cnn}}^{L_{cls\_neg}^{i}}\\
+\frac{\alpha_{2}}{Num} \sum_{i\in Pos} f(\bar{w}_{i}^{m}) \bigtriangledown_{w_{cnn}}^{L_{lreg\_pos}^{i}}
\end{split}
\end{equation}\\
\subsection{Description of the underwater robot picking contest}
The underwater robot picking contest datasets were generated by National Natural Science Foundation of China and Dalian Municipal People's Government. The Chinese website is \url{http://www.cnurpc.org/index.html} and the English website is \url{http://en.cnurpc.org/}

The contest holds annually from 2017, consisting of online and offline object detection contests. In this paper, we use URPC2017 and URPC2018 datasets from the online object detection contest. To use the datasets, participants need to communicate with zhuming@dlut.edu.cn and sign a commitment letter for data usage: \url{http://www.cnurpc.org/a/js/2018/0914/102.html}\\
\subsection{The official leaderboard of the URPC2017 competition}
Table V shows the official leaderboard
of the URPC2017 competition, which is an anonymous leaderboard with mean Average Precision (mAP).
\begin{table}[htbp]\footnotesize
\begin{center}
\caption{The official leaderboard of URPC2017 competition.}
\begin{tabular}{lccccccccccc}
\hline
Method & 1 & 2 & 3 & 4 & 5 & 6 & 7 & 8 & 9\\
mAP(\%) & \textbf{45.1} & 35.7 & 33.4 & 32.0 & 30.2 & 29.6 & 28.8 & 28.4 & 26.6  \\
\hline
\end{tabular}
\end{center}
\label{tab:leadb}
\end{table}

\end{document}